\documentclass[runningheads]{llncs}
\usepackage{adjustbox}
\usepackage[table]{xcolor}
\usepackage{booktabs}
\usepackage{enumitem}
\usepackage{multirow}
\usepackage{amsmath}
\usepackage{amssymb}
\usepackage{hyperref}
\usepackage[T1]{fontenc}
\usepackage{subfig}
\usepackage[dvipsnames]{xcolor}
\usepackage{worldflags}
\usepackage{float}
\usepackage{makecell}
\usepackage{bbm}
\usepackage{graphicx}
\usepackage{subcaption}
\usepackage{pifont}
\usepackage[normalem]{ulem}

\begin{document}
\title{ChronoSurv: A Clinical Pathway-Guided Graph Framework for Multimodal Survival Analysis}
\titlerunning{ChronoSurv}

\author{Hugo Miccinilli\inst{1,*}\and Theo Di Piazza\inst{2,*}}
\authorrunning{Miccinilli and Di Piazza}
\institute{
$^{1}$Université Paris-Saclay, CentraleSupélec, MICS, France \\ 
$^{2}$University of Lyon, INSA Lyon, CREATIS, France\\
$^{*}$Equal contribution\\
\email{\{hugo.miccinilli@centralesupelec.fr, theo.dipiazza@creatis.insa-lyon.fr\} }}

\maketitle              
%

\begin{abstract}
Accurate survival prediction is essential for personalized treatment planning in head and neck cancer, yet remains challenging due to the heterogeneous and high-dimensional nature of multimodal clinical data. While deep survival models have improved predictive performance over classical statistical approaches, existing methods typically rely on static fusion strategies or temporally agnostic modeling, limiting their ability to capture structured clinical workflows.
In this work, we propose ChronoSurv, a heterogeneous hierarchical directed graph framework for multimodal survival analysis. ChronoSurv represents patient care as a progression-aware clinical trajectory using directed graphs aligned with key diagnostic steps. A hierarchical topology incorporates fine-grained, coarse, and global representations, further supporting flexible adaptation to missing modalities, while heterogeneous message passing models complex and asymmetric relationships across modalities and clinical steps.
Experimental results on two public datasets demonstrate that ChronoSurv achieves state-of-the-art discriminative performance while maintaining statistically reliable calibration. Comprehensive ablation studies further confirm the contribution of each architectural component, highlighting the potential of trajectory-aware graph modeling for multimodal survival prediction. Access our code at \url{https://github.com/MICS-Lab/ChronoSurv}.
\end{abstract}

\section{Introduction}
\label{introduction}

The integration of machine learning into precision oncology has significantly advanced survival prediction and treatment planning across a wide range of malignancies~\cite{wang_survey_2022}. Among these, head and neck cancer (HNC) remains one of the most prevalent cancers worldwide~\cite{bray_global_2024}. Despite progress in diagnostic and therapeutic strategies, patient outcomes remain suboptimal, with five-year survival rates still limited~\cite{budach_novel_2019}. A major challenge lies in the heterogeneous and high-dimensional nature of HNC patient data, spanning clinical variables, medical imaging, pathological findings, and reports~\cite{dorrich_multimodal_2024}. Effectively integrating these complementary yet structurally diverse data sources remains an open problem.

Survival analysis aims to predict the time until an event of interest while accounting for censored observations. In HNC, multimodal survival prediction is complicated by substantial discrepancies in scale and granularity across data sources. Bridging these heterogeneous representations, ranging from compact clinical variables to high-dimensional Whole Slide Images (WSIs) and textual reports, requires models that can effectively capture cross-modal interactions while accounting for their distinct structural characteristics.

Early survival analysis methods rely primarily on statistical models~\cite{weibull_statistical_1951,cox_regression_1972}, which are limited in their ability to capture non-linear relationships between patient characteristics and survival risk. Deep learning approaches~\cite{katzman_deepsurv_2018,lee_deephit_2018,hu_transformer-based_2021} have demonstrated improved performance over classical methods~\cite{ishwaran_random_2008,chen_xgboost_2016}, yet most are designed for unimodal inputs.

A prominent line of research in multimodal survival analysis combines WSIs with genomic profiles~\cite{chen2021multimodal,chen2021pan,xiong2024mome,mingcheng_hypergraph_2025,jaume_modeling_2024}, typically extracting modality-specific features followed by feature fusion and survival prediction. While these approaches demonstrate improved performance over unimodal baselines, they often focus on limited modality combinations. In head and neck cancer specifically, prior work has primarily targeted PET/CT-based imaging pipelines~\cite{meng_merging_2023,saeed_survrnc_2024,zhang_improving_2025}, notably around the HECKTOR benchmark~\cite{andrearczyk_overview_2023}. These studies predominantly address segmentation and recurrence-free survival prediction from PET/CT imaging rather than structured fusion across diverse clinical modalities and care stages. More recently, broader multimodal strategies, incorporating tabular data, textual reports, and histopathology, have been explored through late fusion~\cite{vale-silva_long-term_2021}, hierarchical pooling~\cite{li_hfbsurv_2022}, graph-based~\cite{shan_graphmmp_2025} or attention-based~\cite{cui_survival_2022,hou_multimodal_2025} fusion. However, most existing methods treat patient data as an unordered collection of features, overlooking the hierarchical organization of clinical information and the temporal structure of care pathways. In particular, the sequential progression of clinical decision-making, from patient history review through diagnosis to surgical intervention, remains underexplored in current survival frameworks.

Graph Neural Networks (GNNs) provide a principled framework for modeling relational structure in data. In computational pathology, GNNs have been extensively used to capture spatial dependencies between tissue patches within whole slide images~\cite{chen2021whole,zhou2019cgc,lu2021slidegraphslideimagelevel}, organs at risk~\cite{bae_hognet_2024} or to model patient similarity~\cite{kim_gnn-surv_2023}. Their ability to encode heterogeneous, multi-scale relationships makes them a natural candidate for representing structured clinical workflows. To our knowledge, however, no prior work leverages GNNs to explicitly model the chronological progression of clinical care pathways for multimodal survival prediction.

To address these challenges, we propose ChronoSurv, a novel hierarchical directed graph framework for multimodal survival analysis. Inspired by the clinical care workflow, our method explicitly models patient trajectories from background review to diagnosis and surgical intervention, enabling progression-aware prediction. Notably, ChronoSurv integrates a heterogeneous multi-level message passing scheme that jointly captures sub-modality features, clinical steps and patient-level representations while supporting adaptable topology for missing-modality robustness. In a multi-cohort setting combining two public datasets, ChronoSurv demonstrates state-of-the-art discriminative performance alongside reliable calibration, supported by comprehensive ablation studies.

\section{Method}
\label{method}

\begin{figure}[h]
    \centering
    \includegraphics[width=\textwidth]{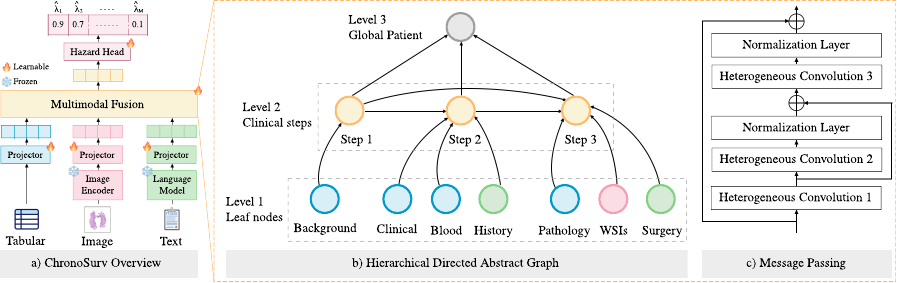}
    \caption{ChronoSurv overview. (a) Multimodal features are integrated through a dedicated fusion module prior to survival prediction. (b) Each patient is modeled as a hierarchical directed graph encoding modality-, step-, and patient-level structure. (c) Heterogeneous convolutions enable cross-level message passing.}
    \label{sph:fig:method_overview}
\end{figure}

As shown in Fig.~\ref{sph:fig:method_overview}, we propose ChronoSurv, a multimodal framework for survival analysis in head and neck cancer that mirrors the clinical care pathway. Our approach extracts per-modality features (Sec.~\ref{sec:method:init}), models each patient as a hierarchical directed graph (Sec.~\ref{sec:method:graph}), enables feature interaction through heterogeneous message passing (Sec.~\ref{sec:method:passing}), and predicts discrete-time hazards (Sec.~\ref{sec:method:head}).

\noindent\textbf{Survival Modeling Setup}. We consider right-censored survival data with patient covariates $X$, event time $T^{*}$ and censoring time $C$. The observed time is $T = \min(T^{*}, C)$ with event indicator $\delta = \mathbbm{1}(T^{*} \leq C)$. We assume conditional non-informative censoring, i.e., $T^{*} \perp\!\!\!\perp C \mid X$. To enable flexible neural modeling of time-to-event outcomes, we adopt a discrete-time survival formulation. Given a prediction horizon $T_{\max}$, we define a time grid $0 = t_0 < t_1 < \cdots < t_M = T_{\max}$. For each interval $j\in \{1, \ldots, M\}$, we model the hazard, denoted as $\lambda(t_j \mid X)$ and defined as $\lambda(t_j \mid X) = \mathbb{P}(T^{*} = t_j \mid T^{*} \geq t_j,\, X)$, from which the conditional survival function follows as $ S(t_j \mid X) = \mathbb{P}(T^{*} > t_j \mid X) = \prod_{k=1}^{j}\bigl(1 - \lambda(t_k \mid X)\bigr)$, enabling computation of survival probabilities and likelihood-based training.

\subsection{Feature Initialization} \label{sec:method:init}
The patient data span a wide range of modalities including WSI ($\mathcal{K}_{\mathrm{wsi}}$), tabular ($\mathcal{K}_{\mathrm{tab}}$), and text ($\mathcal{K}_{\mathrm{text}}$) sub-modalities, with $\mathcal{K} = \mathcal{K}_{\mathrm{tab}} \cup \mathcal{K}_{\mathrm{wsi}} \cup \mathcal{K}_{\mathrm{text}}$. Each sub-modality is projected into a shared $d$-dimensional latent space:
\begin{equation}
    h_{k} =
    f_{k}\left(
        \mathbbm{1}_{k\in\mathcal{K}_{\mathrm{tab}}} x_{k} +
        \mathbbm{1}_{k\in\mathcal{K}_{\mathrm{wsi}}}  \Phi^{\text{visual}}_{\text{enc}}(x_{k}) +
        \mathbbm{1}_{k\in\mathcal{K}_{\mathrm{text}}}  \Phi^{\text{language}}_{\text{model}}(x_{k})
    \right) \, ,
\end{equation}
where $h_{k} \in \mathbb{R}^{d}$ is the encoded representation of sub-modality $k$, each $f_{k}$ is a learnable projector, $\Phi^{\text{visual}}_{\text{enc}}$ a pretrained visual encoder and $\Phi^{\text{language}}_{\text{model}}$ a pretrained language model, both kept frozen. 

\subsection{Graph Construction} \label{sec:method:graph}
Each patient is represented as a hierarchical directed graph $\mathcal{G}=(\mathcal{V},\mathcal{E})$ whose node set is partitioned into three levels (Fig.~\ref{sph:fig:method_overview}b). First-level leaf nodes $\mathcal{V}_{1}$ carry the sub-modality representations $\{h_{k}\}_{k\in\mathcal{K}}$ from Sec.~\ref{sec:method:init}. Second-level nodes $\mathcal{V}_{2}$ encode the temporal structure of the clinical care pathway through three key steps: (1)~patient background, (2)~initial cancer diagnosis, and (3)~local surgery, their features are zero-initialized. Each leaf node is connected to its corresponding clinical step via a directed edge (e.g.\ blood $\to$ step~2, WSIs $\to$ step~3), forming edge set $\mathcal{E}_{1\to2}$. Clinical-step nodes are further linked by directed edges $\mathcal{E}_{\text{temporal}}$ aligned with the sequential ordering of clinical events. The third level is a single patient node $\mathcal{V}_{3}$, initialized by mean-pooling over available leaf embeddings and connected to all step nodes through $\mathcal{E}_{2\to3}$. Node and edge sets are defined as:
\begin{equation}
    \mathcal{V} = \mathcal{V}_{1} \cup \mathcal{V}_{2} \cup
    \mathcal{V}_{3}\,,\qquad
    \mathcal{E} = \mathcal{E}_{1\to2} \cup
    \mathcal{E}_{\text{temporal}} \cup \mathcal{E}_{2\to3}\,.
\end{equation}
Importantly, the graph topology is patient-specific: when a sub-modality $k$ is unavailable, due to differences in clinical protocols, institutional practices, or incomplete records, the corresponding leaf node and its incident edges are omitted from $\mathcal{G}$, providing a principled mechanism for handling missing modalities and making ChronoSurv applicable to heterogeneous real-world clinical settings.

\subsection{Message Passing} \label{sec:method:passing}

ChronoSurv employs three heterogeneous graph attention layers~\cite{brodyHowAttentiveAre2022}, each operating on a distinct relation type to progressively build a patient-level representation. The first layer $\Psi_1$ pools fine-grained sub-modality features into coarser clinical-step representations through $\mathcal{R}_{1\to2} = \{k \to s \mid k \in \mathcal{K}\}$, as follows:
\begin{equation}
    h^{(1)}_{s} = \sum_{r \in \mathcal{R}_{1\to2}}
        \sum_{u \in \mathcal{N}_{r}(s)}
        \Psi_{1}(h_{u},\, h_{s}), \quad s \in \mathcal{V}_{2},
\end{equation}
where $\mathcal{N}_r(s)$ denotes the set of neighbors of $s$ under relation $r$. The second layer $\Psi_2$ captures progression-aware dependencies by propagating information along the clinical trajectory through next ($s_i \!\xrightarrow{\text{N}}\! s_j$, $j{-}i{=}1$) and skip ($s_i \!\xrightarrow{\text{S}}\! s_j$, $j{-}i{>}1$) directed temporal edges between step nodes, such that:
\begin{equation}
    h^{(2)}_{s} = h^{(1)}_{s} + \mathrm{LayerNorm}\!\left(
            \sum_{r \in \mathcal{R}_{\mathrm{t}}}
            \sum_{u \in \mathcal{N}_{r}(s)}
            \Psi_{2}(h^{(1)}_{u},\, h^{(1)}_{s})
        \right), \quad s \in \mathcal{V}_{2}.
\end{equation}
At this stage, step nodes jointly capture aggregated sub-modality information and temporal clinical progression, serving as intermediate representations that bridge fine-grained features and the patient-level embedding. Finally, the third layer $\Psi_3$ fuses all clinical-step representations into a global patient embedding through $\mathcal{R}_{2\to3} = \{s \to g \mid s \in \mathcal{V}_{2}\}$, following:
\begin{equation}
    h_{g} = h_{g}^{(0)} + \mathrm{LayerNorm} \ \!(
            \sum_{r \in \mathcal{R}_{2\to3}}
            \sum_{u \in \mathcal{N}_{r}(g)}
            \Psi_{3}(h^{(2)}_{u},\, h_{g}^{(0)})
        ),
\end{equation}
where $h_g^{(0)}$ denotes the mean-pooled initialization of the patient node.

\subsection{Survival Head} \label{sec:method:head}
The survival head receives the patient embedding $h_g$ and predicts bin-wise hazards over the discrete grid defined as follows:
\begin{equation}
    \hat{\boldsymbol{\lambda}}
    = \sigma\!\left(
        \Psi^{\text{hazard}}_{\text{head}}(h_{g})
    \right) \in [0,1]^{M},
\end{equation}
where $\hat{\lambda}_{j}$ estimates the conditional event
probability in bin $j$ given survival up to bin $j{-}1$.
Let $\kappa(t_i)$ denote the bin index of observed time $t_i$
and $y_{ik} := \mathbbm{1}(k = \kappa(t_i),\, \delta_i = 1)$.
Training minimizes
$\mathcal{L} = \mathcal{L}_{\text{NLL}}
+ \beta \, \mathcal{L}_{\text{X-CAL}}$, where:
\begin{equation}
    \mathcal{L}_{\text{NLL}}
    = -\frac{1}{N} \sum_{i=1}^{N}
      \sum_{k=1}^{\kappa(t_i)} \Bigl[
        y_{ik} \log \hat{\lambda}_{ik}
        + (1 - y_{ik}) \log(1 - \hat{\lambda}_{ik})
      \Bigr],
\end{equation}
is the discrete-time negative log-likelihood~\cite{brownUseIndicatorVariables1975}, $\mathcal{L}_{\text{X-CAL}}$ a differentiable calibration penalty~\cite{goldsteinXCALExplicitCalibration2020}, and $\beta$ a tunable weighting coefficient. Day-level survival predictions are obtained by assigning each day to its corresponding time bin.

\section{Experiments}
\label{experimental}

\noindent \textbf{Datasets.} We leverage two public head and neck
cancer datasets for  multimodal survival analysis: \texttt{HANCOCK}~\cite{dorrich_multimodal_2024} and \texttt{TCGA-HNSC}~\cite{lawrence_comprehensive_2015}, summarized in Table~\ref{tab:datasets}.
\texttt{HANCOCK} provides clinical, blood, and pathological data, textual surgical reports, and WSIs from both primary tumors and lymph nodes. \texttt{TCGA-HNSC} covers clinical and pathological data with tumor WSIs only, lacking blood data, textual reports, and
lymph-node slides. We further construct \texttt{UniHNC} by combining both sources, creating a multi-cohort setting that reflects real-world clinical heterogeneity where modality availability varies across institutions (Sec.~\ref{sec:method:graph}).

\newcommand{\cmark}{\ding{51}}%
\newcommand{\xmark}{\ding{55}}%

\begin{table}[h]
\centering
\footnotesize
\setlength{\tabcolsep}{5pt}
\caption{Dataset overview.
\cmark / \xmark\, indicate modality availability.}
\begin{tabular}{l l c c c c c}
\toprule
Dataset & Origin & Samples & Censoring
& Tabular & WSI & Text \\
\midrule
\texttt{HANCOCK}~\cite{dorrich_multimodal_2024}   & Germany & 763 & 72.1\,\%
& \cmark & \cmark & \cmark \\
\texttt{TCGA-HNSC}~\cite{lawrence_comprehensive_2015} & USA & 526 & 57.6\,\%
& \cmark & \cmark & \xmark \\
\midrule
\texttt{UniHNC} & Germany \& USA & 1{,}289 & 66.1\,\% 
& \cmark & \cmark & \cmark \\
\bottomrule
\end{tabular}
\label{tab:datasets}
\end{table}

\noindent \textbf{Setup.}
Experiments use 5-fold cross-validation with patient-level 70/15/15 splits. Models are trained with early stopping on the validation set loss and evaluated on the held-out test fold, reporting mean and standard deviation across folds.
All models leverage the same frozen feature extraction module: UNI~\cite{chen_towards_2024} as visual encoder and BioClinicalBERT~\cite{alsentzer_publicly_2019} as language model. ChronoSurv is trained for 20{,}000 iterations with a batch size of 8 using AdamW, a $3{\times}10^{-4}$ learning rate, a $5{\times}10^{-4}$ weight decay and a cosine annealing scheduler. The number of time bins is set to $M=100$ (tuned over $\{10, 50, 100, 500, 1000\}$), the calibration weight to $\beta=7$ (tuned over $\{0.1, 1, 3, 5, 7, 15\}$) and the embedding dimension to $d=512$. Inference takes approximately 30\,ms per sample on a single RTX 2080 Ti.

\subsection{Evaluation results}
\label{experimental:quantitative}

Table~\ref{tab:quantitative_survival} compares ChronoSurv against \textit{Non-Deep Learning} ({\color{Cerulean}\small$\blacksquare$}), \textit{Deep Learning} ({\color{Orchid}\small$\blacksquare$}), and \textit{Multimodal Deep Learning} ({\color{Orchid}\small$\blacklozenge$}) baselines. Non-Deep Learning and Deep Learning baselines receive the concatenation of all features  and missing sub-modalities are zero-imputed for all baselines. Discrimination performance is evaluated using C$_{\text{index}}$~\cite{uno_2011}, while the IBS~\cite{graf_assessment_1999} assesses overall survival probability accuracy.
ChronoSurv achieves the highest C$_{\text{index}}$ on both \texttt{UniHNC} and \texttt{HANCOCK} while obtaining the best or tied-best IBS across all settings. Among multimodal baselines, SurvPCG reaches the closest C$_{\text{index}}$ on \texttt{UniHNC} but with a higher IBS, suggesting that although cross-modal attention captures relevant interactions for risk ranking, it yields less
accurate survival probability estimates. On \texttt{TCGA-HNSC}, where text and blood modalities are  unavailable, ChronoSurv matches the best IBS while maintaining competitive discrimination performance, including against MMD, which is designed to handle incomplete multimodal data. Importantly, when applying the D-Calibration test~\cite{haider2018effectivewaysbuildevaluate} with a Type I error rate  of $\alpha = 0.05$, ChronoSurv passes the test in all 5 folds ($p \geq 0.05$), whereas SurvPCG passes only 1/5. Baselines with consistent calibration (Cox, RSF, MultiSurv: 5/5) also achieve competitive IBS but substantially lower discrimination.

\begin{table}[t]
\centering
\footnotesize
\caption{
Performance on the \texttt{UniHNC} dataset, and $^{(\dag)}$
dataset-specific results on \texttt{HANCOCK} and
\texttt{TCGA-HNSC}.
\textbf{Best} in bold, \underline{second best} underlined.
\newline
{\color{Cerulean}\small$\blacksquare$} Non-Deep Learning
{\color{Orchid}\small$\blacksquare$} Deep Learning
{\color{Orchid}\small$\blacklozenge$} Multimodal Deep Learning.
}
\setlength{\tabcolsep}{7pt}
\begin{adjustbox}{width=\columnwidth,center}
\begin{tabular}{l c c c c c c}
\toprule

& \multicolumn{2}{c}{\texttt{UniHNC}} & \multicolumn{2}{c}{\texttt{HANCOCK}$^{(\dag)}$} & \multicolumn{2}{c}{\texttt{TCGA-HNSC}$^{(\dag)}$} \\

\cmidrule(lr){2-3} \cmidrule(lr){4-5} \cmidrule(lr){6-7}

Method & 
C$_{\text{index}}$ ($\uparrow$) & 
IBS ($\downarrow$) & 

C$_{\text{index}}$ ($\uparrow$) & 
IBS ($\downarrow$) & 

C$_{\text{index}}$ ($\uparrow$) & 
IBS ($\downarrow$) \\


\midrule

{\color{Cerulean}\small$\blacksquare$} Weibull~\cite{weibull_statistical_1951} & 
.636 \scriptsize \textcolor{gray}{$\pm$.036} &  
.167 \scriptsize \textcolor{gray}{$\pm$.018} &  
.589 \scriptsize \textcolor{gray}{$\pm$.059} &  
.150 \scriptsize \textcolor{gray}{$\pm$.016} &  
.593 \scriptsize \textcolor{gray}{$\pm$.065} &  
.195 \scriptsize \textcolor{gray}{$\pm$.024} \\  

{\color{Cerulean}\small$\blacksquare$} CoxPH~\cite{cox_regression_1972} & 
.652 \scriptsize \textcolor{gray}{$\pm$.032} &  
.155 \scriptsize \textcolor{gray}{$\pm$.008} &  
.623 \scriptsize \textcolor{gray}{$\pm$.035} &  
.138 \scriptsize \textcolor{gray}{$\pm$.012} &  
.619 \scriptsize \textcolor{gray}{$\pm$.059} &  
\underline{.183} \scriptsize \textcolor{gray}{$\pm$.008} \\  

{\color{Cerulean}\small$\blacksquare$} RSF~\cite{ishwaran_random_2008} & 
.654 \scriptsize \textcolor{gray}{$\pm$.032} &  
\underline{.154} \scriptsize \textcolor{gray}{$\pm$.012} &  
.622 \scriptsize \textcolor{gray}{$\pm$.019} &  
\underline{.136} \scriptsize \textcolor{gray}{$\pm$.009} &  
.584 \scriptsize \textcolor{gray}{$\pm$.080} &  
\underline{.183} \scriptsize \textcolor{gray}{$\pm$.018} \\  


{\color{Orchid}\small$\blacksquare$} DeepSurv~\cite{katzman_deepsurv_2018} & 
.674 \scriptsize \textcolor{gray}{$\pm$.028} &  
.165 \scriptsize \textcolor{gray}{$\pm$.013} &  
.644 \scriptsize \textcolor{gray}{$\pm$.048} &  
.149 \scriptsize \textcolor{gray}{$\pm$.022} &  
.639 \scriptsize \textcolor{gray}{$\pm$.068} &  
.191 \scriptsize \textcolor{gray}{$\pm$.015} \\  

{\color{Orchid}\small$\blacksquare$} DeepHit~\cite{lee_deephit_2018} & 
.619 \scriptsize \textcolor{gray}{$\pm$.030} &  
.178 \scriptsize \textcolor{gray}{$\pm$.006} &  
.516 \scriptsize \textcolor{gray}{$\pm$.013} &  
.143 \scriptsize \textcolor{gray}{$\pm$.014} &  
.619 \scriptsize \textcolor{gray}{$\pm$.058} &  
.231 \scriptsize \textcolor{gray}{$\pm$.020} \\  

{\color{Orchid}\small$\blacksquare$} TransDSA~\cite{hu_transformer-based_2021} & 
.629 \scriptsize \textcolor{gray}{$\pm$.024} &  
.263 \scriptsize \textcolor{gray}{$\pm$.045} &  
.602 \scriptsize \textcolor{gray}{$\pm$.027} &  
.255 \scriptsize \textcolor{gray}{$\pm$.077} &  
.514 \scriptsize \textcolor{gray}{$\pm$.056} &  
.280 \scriptsize \textcolor{gray}{$\pm$.018} \\  

{\color{Orchid}\small$\blacklozenge$} MMD~\cite{cui_survival_2022} & 
.683 \scriptsize \textcolor{gray}{$\pm$.024} &  
.184 \scriptsize \textcolor{gray}{$\pm$.028} &  
.659 \scriptsize \textcolor{gray}{$\pm$.031} &  
.187 \scriptsize \textcolor{gray}{$\pm$.025} &  
\underline{.659} \scriptsize \textcolor{gray}{$\pm$.067} &  
\textbf{.182} \scriptsize \textcolor{gray}{$\pm$.035} \\  

{\color{Orchid}\small$\blacklozenge$} GraphMMP~\cite{shan_graphmmp_2025} & 
.667 \scriptsize \textcolor{gray}{$\pm$.038} &  
.161 \scriptsize \textcolor{gray}{$\pm$.026} &  
.657 \scriptsize \textcolor{gray}{$\pm$.019} &  
.145 \scriptsize \textcolor{gray}{$\pm$.014} &  
.633 \scriptsize \textcolor{gray}{$\pm$.073} &  
.187 \scriptsize \textcolor{gray}{$\pm$.045} \\  

{\color{Orchid}\small$\blacklozenge$} HFBSurv~\cite{li_hfbsurv_2022} & 
.679 \scriptsize \textcolor{gray}{$\pm$.019} &  
.302 \scriptsize \textcolor{gray}{$\pm$.030} &  
.656 \scriptsize \textcolor{gray}{$\pm$.042} &  
.349 \scriptsize \textcolor{gray}{$\pm$.063} &  
\textbf{.660} \scriptsize \textcolor{gray}{$\pm$.067} &  
.238 \scriptsize \textcolor{gray}{$\pm$.036} \\  

{\color{Orchid}\small$\blacklozenge$} MultiSurv~\cite{vale-silva_long-term_2021} & 
.651 \scriptsize \textcolor{gray}{$\pm$.045} &  
.156 \scriptsize \textcolor{gray}{$\pm$.009} &  
.637 \scriptsize \textcolor{gray}{$\pm$.041} &  
.141 \scriptsize \textcolor{gray}{$\pm$.014} &  
.591 \scriptsize \textcolor{gray}{$\pm$.058} &  
\textbf{.182 \scriptsize \textcolor{gray}{$\pm$.006}} \\  

{\color{Orchid}\small$\blacklozenge$} SurvPCG~\cite{hou_multimodal_2025} & 
\underline{.695} \scriptsize \textcolor{gray}{$\pm$.023} &  
.163 \scriptsize \textcolor{gray}{$\pm$.012} &  
\underline{.672} \scriptsize \textcolor{gray}{$\pm$.038} &  
.140 \scriptsize \textcolor{gray}{$\pm$.011} &  
\underline{.659} \scriptsize \textcolor{gray}{$\pm$.060} &  
.198 \scriptsize \textcolor{gray}{$\pm$.029} \\  

\cellcolor{blue!5}{\color{Orchid}$\blacklozenge$} ChronoSurv (Ours) & 
\cellcolor{blue!5}\textbf{.702} \scriptsize \textcolor{gray}{$\pm$.039} &  
\cellcolor{blue!5}\textbf{.153} \scriptsize \textcolor{gray}{$\pm$.019} &  
\cellcolor{blue!5}\textbf{.704} \scriptsize \textcolor{gray}{$\pm$.037} &  
\cellcolor{blue!5}\textbf{.134} \scriptsize \textcolor{gray}{$\pm$.020} &  
\cellcolor{blue!5}.652 \scriptsize \textcolor{gray}{$\pm$.048} &  
\cellcolor{blue!5}\textbf{.182} \scriptsize \textcolor{gray}{$\pm$.022} \\  

\bottomrule
\end{tabular}
\end{adjustbox}
\label{tab:quantitative_survival}
\end{table}

\subsection{Ablation study}
\label{experimenta:ablation}

\noindent \textbf{Effect of multimodal fusion.} Table~\ref{tab:ablation_study_aggregation} compares our multimodal fusion module with unimodal baselines ({\color{Lavender}\small$\blacksquare$}) and variants replacing the fusion module by widely used aggregation mechanisms ({\color{LimeGreen}\small$\blacksquare$}), while keeping identical feature initialization and survival heads. Unimodal baselines confirm that each modality carries prognostic signal. In fact, tabular features alone match or surpass several multimodal strategies, highlighting that naïve fusion can be detrimental. In contrast, ChronoSurv consistently outperforms all alternative aggregation schemes, supporting the effectiveness of structured multimodal interaction modeling.


\begin{table}[t]
\centering
\footnotesize
\caption{
Comparison with unimodal baselines and alternative multimodal aggregation schemes.
{\color{Lavender}\small$\blacksquare$} Unimodal. 
{\color{LimeGreen}\small$\blacksquare$} Multimodal.
} 
\setlength{\tabcolsep}{4pt}
\begin{adjustbox}{width=\columnwidth,center}
\begin{tabular}{l c c c c c c}
\toprule
& \multicolumn{2}{c}{\texttt{UniHNC}} & \multicolumn{2}{c}{\texttt{HANCOCK}$^{(\dag)}$} & \multicolumn{2}{c}{\texttt{TCGA-HNSC}$^{(\dag)}$} \\

\cmidrule(lr){2-3} \cmidrule(lr){4-5} \cmidrule(lr){6-7}

Method & 
C$_{\text{index}}$ ($\uparrow$) & 
IBS ($\downarrow$) & 

C$_{\text{index}}$ ($\uparrow$) & 
IBS ($\downarrow$) & 

C$_{\text{index}}$ ($\uparrow$) & 
IBS ($\downarrow$) \\

\midrule

{\color{Lavender}\small$\blacksquare$} Text-only &
.606 \scriptsize \textcolor{gray}{$\pm$.037} &  
\underline{.160} \scriptsize \textcolor{gray}{$\pm$.011} &  
.543 \scriptsize \textcolor{gray}{$\pm$.036} &  
.143 \scriptsize \textcolor{gray}{$\pm$.017} &  
- &  
- \\ 

{\color{Lavender}\small$\blacksquare$} Image-only &
.638 \scriptsize \textcolor{gray}{$\pm$.012} &  
.172 \scriptsize \textcolor{gray}{$\pm$.024} &  
.652 \scriptsize \textcolor{gray}{$\pm$.044} &  
.147 \scriptsize \textcolor{gray}{$\pm$.020} &  
.549 \scriptsize \textcolor{gray}{$\pm$.055} &  
\underline{.212} \scriptsize \textcolor{gray}{$\pm$.033} \\ 

{\color{Lavender}\small$\blacksquare$} Tabular-only &
.668 \scriptsize \textcolor{gray}{$\pm$.065} &  
.170 \scriptsize \textcolor{gray}{$\pm$.039} &  
.665 \scriptsize \textcolor{gray}{$\pm$.090} &  
\underline{.138} \scriptsize \textcolor{gray}{$\pm$.021} &  
.596 \scriptsize \textcolor{gray}{$\pm$.044} &  
.219 \scriptsize \textcolor{gray}{$\pm$.076} \\ 

\midrule

{\color{LimeGreen}\small$\blacksquare$} Mean pooling &
.641 \scriptsize \textcolor{gray}{$\pm$.056} &  
.228 \scriptsize \textcolor{gray}{$\pm$.043} &  
.643 \scriptsize \textcolor{gray}{$\pm$.061} &  
.222 \scriptsize \textcolor{gray}{$\pm$.041} &  
.522 \scriptsize \textcolor{gray}{$\pm$.102} &  
.239 \scriptsize \textcolor{gray}{$\pm$.066} \\ 

{\color{LimeGreen}\small$\blacksquare$} MLP &
.655 \scriptsize \textcolor{gray}{$\pm$.063} &  
.214 \scriptsize \textcolor{gray}{$\pm$.043} &  
.638 \scriptsize \textcolor{gray}{$\pm$.059} &  
.213 \scriptsize \textcolor{gray}{$\pm$.035} &  
.579 \scriptsize \textcolor{gray}{$\pm$.114} &  
.222 \scriptsize \textcolor{gray}{$\pm$.070} \\ 

{\color{LimeGreen}\small$\blacksquare$} Gated Fusion &
.624 \scriptsize \textcolor{gray}{$\pm$.066} &  
.226 \scriptsize \textcolor{gray}{$\pm$.043} &  
.592 \scriptsize \textcolor{gray}{$\pm$.063} &  
.224 \scriptsize \textcolor{gray}{$\pm$.044} &  
.543 \scriptsize \textcolor{gray}{$\pm$.127} &  
.238 \scriptsize \textcolor{gray}{$\pm$.065} \\ 

{\color{LimeGreen}\small$\blacksquare$} FC GNN &
.674 \scriptsize \textcolor{gray}{$\pm$.040} &  
.216 \scriptsize \textcolor{gray}{$\pm$.051} &  
.672 \scriptsize \textcolor{gray}{$\pm$.047} &  
.206 \scriptsize \textcolor{gray}{$\pm$.050} &  
.598 \scriptsize \textcolor{gray}{$\pm$.052} &  
.233 \scriptsize \textcolor{gray}{$\pm$.068} \\ 

{\color{LimeGreen}\small$\blacksquare$} Cross-attention &
.689 \scriptsize \textcolor{gray}{$\pm$.031} &  
.235 \scriptsize \textcolor{gray}{$\pm$.053} &  
.677 \scriptsize \textcolor{gray}{$\pm$.043} &  
.227 \scriptsize \textcolor{gray}{$\pm$.044} &  
\textbf{.656} \scriptsize \textcolor{gray}{$\pm$.052} &  
.245 \scriptsize \textcolor{gray}{$\pm$.092} \\ 

{\color{LimeGreen}\small$\blacksquare$} Self-attention &
\underline{.695} \scriptsize \textcolor{gray}{$\pm$.038} &  
.213 \scriptsize \textcolor{gray}{$\pm$.055} &  
\underline{.697} \scriptsize \textcolor{gray}{$\pm$.039} &  
.200 \scriptsize \textcolor{gray}{$\pm$.050} &  
.644 \scriptsize \textcolor{gray}{$\pm$.025} &  
.236 \scriptsize \textcolor{gray}{$\pm$.076} \\ 

\cellcolor{blue!5}{\color{LimeGreen}\small$\blacksquare$} ChronoSurv (Ours) & 
\cellcolor{blue!5}\textbf{.702} \scriptsize \textcolor{gray}{$\pm$.039} &   
\cellcolor{blue!5}\textbf{.153} \scriptsize \textcolor{gray}{$\pm$.019} &   
\cellcolor{blue!5}\textbf{.704} \scriptsize \textcolor{gray}{$\pm$.037} &   
\cellcolor{blue!5}\textbf{.134} \scriptsize \textcolor{gray}{$\pm$.020} &   
\cellcolor{blue!5}\underline{.652} \scriptsize \textcolor{gray}{$\pm$.048} &   
\cellcolor{blue!5}\textbf{.182} \scriptsize \textcolor{gray}{$\pm$.022} \\  

\bottomrule[1.2pt]

\end{tabular}
\end{adjustbox}
\label{tab:ablation_study_aggregation}
\end{table}

\noindent \textbf{Component-wise ablation.} Table~\ref{tab:ablation_study_component} reports a leave-one-out ablation study on ChronoSurv's components. Removing any individual clinical step and its corresponding leaf nodes (1-3) degrades performance across all datasets, highlighting that each clinical stage provides complementary prognostic information. Notably, excluding step 2 (initial cancer diagnosis) yields the largest reduction in C$_{\text{index}}$. Removing hierarchical levels (4 \& 5) results in a performance decline, indicating that incorporating coarser information captured by intermediate clinical steps and global patient representations is critical for effective multimodal integration. Disabling heterogeneous message passing (6) consistently degrades C$_{\text{index}}$ and IBS across datasets, demonstrating the importance of relation-specific modeling for structured, multi-level message passing. Finally, replacing directed edges with undirected ones (7) also degrades performance, suggesting that preserving the temporally ordered topology leads to more accurate survival prediction.

\begin{table}[t]
\centering
\footnotesize
\setlength{\tabcolsep}{4pt}
\caption{
Leave-one-out ablation on clinical-steps modeling (1-3), hierarchical levels (4 and 5), graph heterogeneity (6) and edge directionality (7).
} 
\begin{adjustbox}{width=\columnwidth,center}
\begin{tabular}{l c c c c c c}
\toprule
& \multicolumn{2}{c}{\texttt{UniHNC}} & \multicolumn{2}{c}{\texttt{HANCOCK}$^{(\dag)}$} & \multicolumn{2}{c}{\texttt{TCGA-HNSC}$^{(\dag)}$} \\

\cmidrule(lr){2-3} \cmidrule(lr){4-5} \cmidrule(lr){6-7}

Method & 
C$_{\text{index}}$ ($\uparrow$) & 
IBS ($\downarrow$) & 

C$_{\text{index}}$ ($\uparrow$) & 
IBS ($\downarrow$) & 

C$_{\text{index}}$ ($\uparrow$) & 
IBS ($\downarrow$) \\

\midrule

\cellcolor{blue!5}ChronoSurv & 
\cellcolor{blue!5}\textbf{.702} \scriptsize \textcolor{gray}{$\pm$.039} &   
\cellcolor{blue!5}\textbf{.153} \scriptsize \textcolor{gray}{$\pm$.019} &   
\cellcolor{blue!5}\textbf{.704} \scriptsize \textcolor{gray}{$\pm$.037} &   
\cellcolor{blue!5}\textbf{.134} \scriptsize \textcolor{gray}{$\pm$.020} &   
\cellcolor{blue!5}\textbf{.652} \scriptsize \textcolor{gray}{$\pm$.048} &   
\cellcolor{blue!5}.182 \scriptsize \textcolor{gray}{$\pm$.022} \\  

\toprule

(1) w/o step 1 &
.644 \scriptsize \textcolor{gray}{$\pm$.088} &  
.162 \scriptsize \textcolor{gray}{$\pm$.016} &  
.638 \scriptsize \textcolor{gray}{$\pm$.045} &  
.143 \scriptsize \textcolor{gray}{$\pm$.018} &  
.596 \scriptsize \textcolor{gray}{$\pm$.112} &  
.193 \scriptsize \textcolor{gray}{$\pm$.029} \\  

(2) w/o step 2 &
.586 \scriptsize \textcolor{gray}{$\pm$.056} &  
.182 \scriptsize \textcolor{gray}{$\pm$.016} &  
.552 \scriptsize \textcolor{gray}{$\pm$.058} &  
.145 \scriptsize \textcolor{gray}{$\pm$.011} &  
.586 \scriptsize \textcolor{gray}{$\pm$.026} &  
.239 \scriptsize \textcolor{gray}{$\pm$.036} \\  

(3) w/o step 3 &
.607 \scriptsize \textcolor{gray}{$\pm$.050} &  
.171 \scriptsize \textcolor{gray}{$\pm$.021} &  
.607 \scriptsize \textcolor{gray}{$\pm$.070} &  
.141 \scriptsize \textcolor{gray}{$\pm$.021} &  
.564 \scriptsize \textcolor{gray}{$\pm$.047} &  
.217 \scriptsize \textcolor{gray}{$\pm$.031} \\  

\midrule

(4) w/o level 2 &
.546 \scriptsize \textcolor{gray}{$\pm$.061} &  
.186 \scriptsize \textcolor{gray}{$\pm$.018} &  
.546 \scriptsize \textcolor{gray}{$\pm$.066} &  
.147 \scriptsize \textcolor{gray}{$\pm$.019} &  
.529 \scriptsize \textcolor{gray}{$\pm$.088} &  
.246 \scriptsize \textcolor{gray}{$\pm$.024} \\  

(5) w/o level 3 &
.598 \scriptsize \textcolor{gray}{$\pm$.067} &  
.181 \scriptsize \textcolor{gray}{$\pm$.022} &  
.579 \scriptsize \textcolor{gray}{$\pm$.052} &  
.149 \scriptsize \textcolor{gray}{$\pm$.018} &  
.552 \scriptsize \textcolor{gray}{$\pm$.056} &  
.231 \scriptsize \textcolor{gray}{$\pm$.043} \\  

\midrule

(6) w/o heterogeneity &
.644 \scriptsize \textcolor{gray}{$\pm$.043} &  
.157 \scriptsize \textcolor{gray}{$\pm$.009} &  
.598 \scriptsize \textcolor{gray}{$\pm$.062} &  
.144 \scriptsize \textcolor{gray}{$\pm$.012} &  
.631 \scriptsize \textcolor{gray}{$\pm$.047} &  
\textbf{.179} \scriptsize \textcolor{gray}{$\pm$.018} \\  

(7) w/o directionality &
.681 \scriptsize \textcolor{gray}{$\pm$.058} &  
.199 \scriptsize \textcolor{gray}{$\pm$.039} &  
.679 \scriptsize \textcolor{gray}{$\pm$.031} &  
.196 \scriptsize \textcolor{gray}{$\pm$.035} &  
.646 \scriptsize \textcolor{gray}{$\pm$.054} &  
.211 \scriptsize \textcolor{gray}{$\pm$.049} \\ 

\bottomrule[1.2pt]

\end{tabular}
\end{adjustbox}
\label{tab:ablation_study_component}
\end{table}

\subsection{Qualitative results} \label{sec:experiments:qualitative}

\begin{figure}[t]
    \centering
    \includegraphics[width=\textwidth]{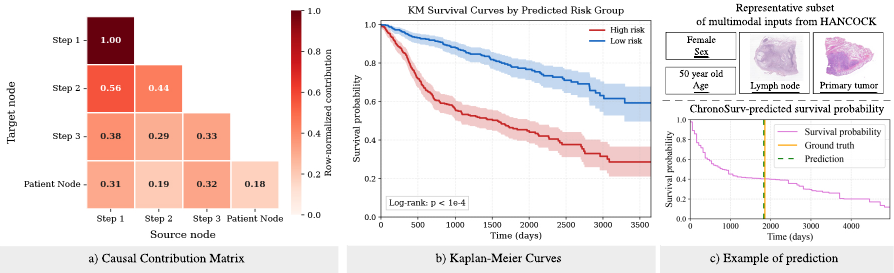}
    \caption{Qualitative results. (a) Contribution matrix between clinical steps and the global patient node, illustrating the row-normalized magnitude of message-passing contributions from each source node to its corresponding target. (b) Kaplan-Meier survival curves stratified by predicted risk groups. (c) Patient-level prediction displaying the estimated survival curve and survival time.}
    \label{sph:fig:qualitative}
\end{figure}

Fig.~\ref{sph:fig:qualitative} provides interpretability insights into ChronoSurv's learned representations. The contribution matrix (a) indicates how information flows through the graph hierarchy, extracted from normalized contribution between two nodes. These results indicate that step 1 (background) acts as a self-contained source, step 2 (initial diagnosis) predominantly aggregates background patient information from step 1, while step 3 (local surgery) draws from all preceding steps, further highlighting the complementary nature of clinical steps. At the patient level, step 1 and step 3 contribute most to the global representation, reflecting the clinical importance of both patient history and surgical findings for prognosis. Kaplan–Meier analysis (b) demonstrates clear risk stratification, showing a statistically significant difference between groups (log-rank test $p < 10^{-4}$).

\section{Discussion and Conclusion}
\label{conclusion}

We introduce ChronoSurv, a heterogeneous hierarchical directed graph framework for multimodal survival analysis in head and neck cancer. By modeling the clinical care pathway as a temporally directed graph with heterogeneous message passing, ChronoSurv captures progression-aware dependencies while naturally handling missing modalities through adaptive topology. Experiments on two public datasets show that structured, progression-aware graph modeling yields both accurate risk ranking and well-calibrated survival estimates, highlighting the potential of clinically-guided modeling for multimodal survival analysis.

\noindent \textbf{Limitations and Future work.} Although our study primarily focuses on multimodal fusion, we anticipate that opportunities for improvement could come from fine-tuning initial feature extractors or exploring alternative visual encoders and language models, which is left for future work. Additionally, future work will explore scaling the framework to larger cohorts and more diverse malignancies as additional annotations become available.

\subsubsection{\discintname}
The authors have no competing interests to declare that are relevant to the content of this article.

\bibliographystyle{splncs04}
\bibliography{Paper-3722}

\end{document}